\documentclass[10pt,twocolumn,letterpaper]{article}

\usepackage{ijcb}
\usepackage{times}
\usepackage{epsfig}
\usepackage{graphicx}
\usepackage{amsmath}
\usepackage{amssymb}
\usepackage[norule,symbol,perpage]{footmisc}

\usepackage{booktabs}
\usepackage{multirow}
\usepackage{subfigure}
\usepackage{url}

\ijcbfinalcopy 

\ifijcbfinal\fi

\makeatletter
\def\ps@IEEEtitlepagestyle{
\def\@oddfoot{\mycopyrightnotice}
\def\@evenfoot{}
}
\def\mycopyrightnotice{
{\hfill \footnotesize 978-1-7281-9186-7/20/\$31.00 \copyright 2020 IEEE\hfill}
}
\makeatother

\begin{document}
\title{Open Source Iris Recognition Hardware and Software \\with Presentation Attack Detection}

\author{Zhaoyuan Fang\\
Department of Electrical Engineering\\University of Notre Dame\\
Notre Dame, IN 46556\\
{\tt\small zfang@nd.edu}
\and
Adam Czajka\\
Department of Computer Science\\University of Notre Dame\\
Notre Dame, IN 46556\\
{\tt\small aczajka@nd.edu}
}

\maketitle
\thispagestyle{empty}

\begin{abstract}
This paper proposes the first known to us open source hardware and software iris recognition system with presentation attack detection (PAD), which can be easily assembled for about 75 USD using Raspberry Pi board and a few peripherals. The primary goal of this work is to offer a low-cost baseline for spoof-resistant iris recognition, which may (a) stimulate research in iris PAD and allow for easy prototyping of secure iris recognition systems, (b) offer a low-cost secure iris recognition alternative to more sophisticated systems, and (c) serve as an educational platform. We propose a lightweight image complexity-guided convolutional network for fast and accurate iris segmentation, domain-specific human-inspired Binarized Statistical Image Features (BSIF) to build an iris template, and to combine 2D (iris texture) and 3D (photometric stereo-based) features for PAD. The proposed iris recognition runs in about 3.2 seconds and the proposed PAD runs in about 4.5 seconds on Raspberry Pi 3B+. The hardware specifications and all source codes of the entire pipeline are made available along with this paper.
\end{abstract}


\begin{figure*}[!htb]
\centering
\subfigure[]{\includegraphics[width=0.253\textwidth]{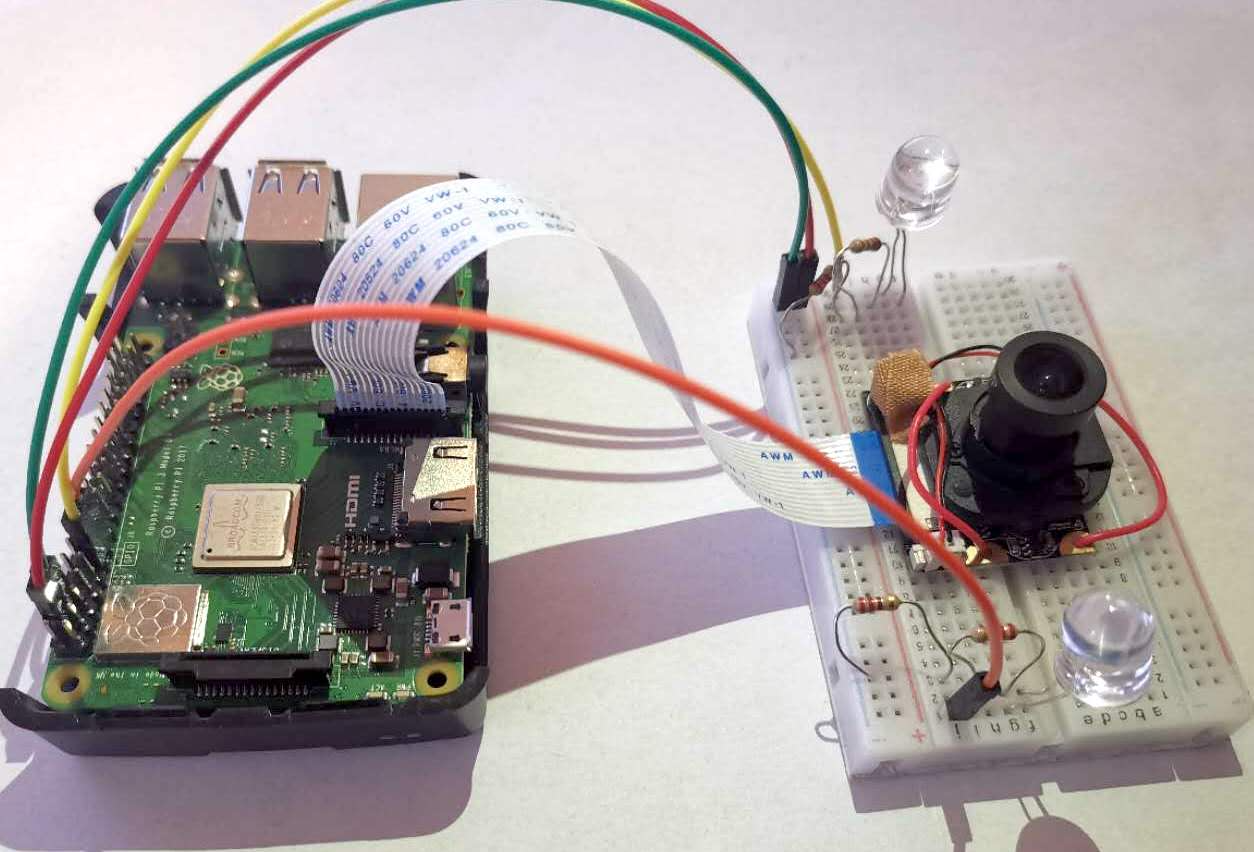}}\hfill
\subfigure[]{\includegraphics[width=0.26\textwidth]{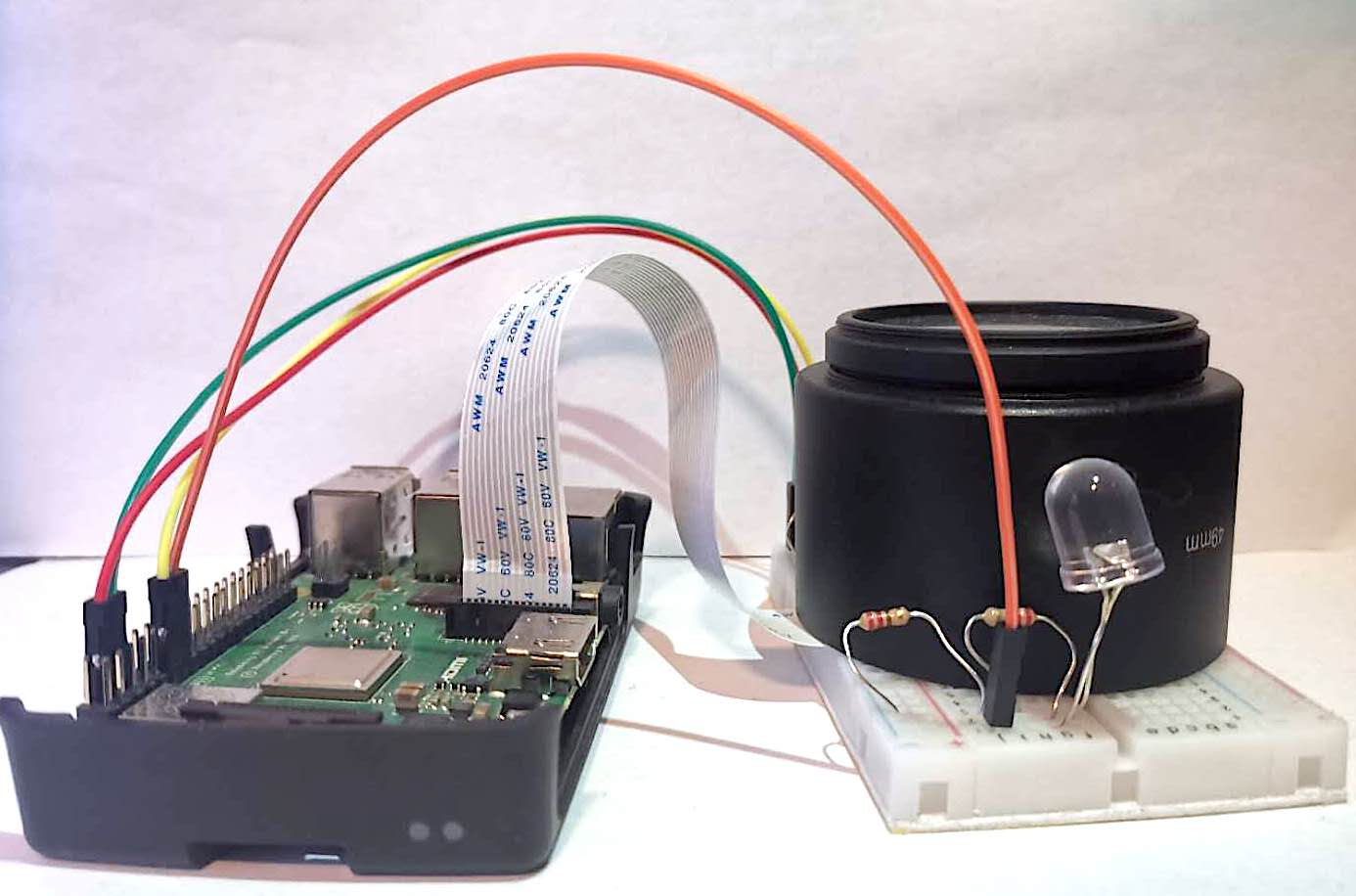}}\hfill
\subfigure[]{\includegraphics[width=0.23\textwidth]{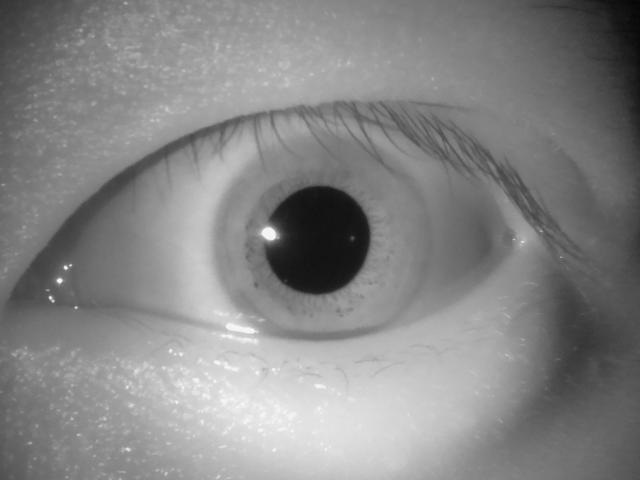}}\hfill
\subfigure[]{\includegraphics[width=0.23\textwidth]{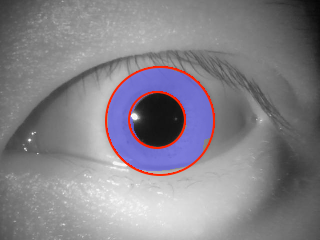}}
\caption{Prototype of the proposed open source iris recognition system without (a) and with (b) metal sleeve and near-infrared filter covering the lens, along with an example image taken by this prototype (c) and segmentation results (d) generated by the proposed method (mask prediction is shown in blue and circular approximations of iris and pupil boundaries are shown in red).}
\label{fig:setup}
\end{figure*}

\section{Introduction}

Iris recognition is without a doubt one of the main biometric modes, and the seminal work by Daugman~\cite{Daugman_1993_recognition} paved the way for many current commercial systems and inspired academia with a mathematically elegant and information theory-based iris feature extraction and matching. ISO/IEC 19794-6 recommendations regarding the iris image properties are not demanding in terms of resolution (640$\times$480 pixels with at least 120 pixels per iris diameter), and near-infrared (NIR) LEDs with a peak in the ISO-recommended range of 700-900 nm are easily available. However, very few works have been published along the line of open source iris recognition~\cite{Lee2009VASIR, Masek2003MATLABSC, Othman_2016_OSIRIS}, and all previous papers focused on the software domain. To the best of our knowledge, there has not been prior work that attempted to offer an open source hardware design for iris recognition.\par


In this paper, we propose an open source iris recognition system (hardware and software) with presentation attack detection (PAD). The hardware platform is based on the Raspberry Pi 3B+ board, through which we perform illumination control, image acquisition, and the entire image processing involved in the pipeline. In addition to the main Pi board, the hardware involves a NIR-sensitive Pi camera, an NIR-filter, two NIR LEDs, a breadboard, resistors, and jumper wires. We designed our hardware in a way such that it is low-cost, easy to assemble, easy to use, and follows the ISO recommendations regarding the iris image capture (thus the use of NIR, rather than visible light). The total cost of the hardware in the moment of writing this paper was less than 75 USD, which was cheaper than most commercial iris systems. Our open-source software pipeline consists of three modules: segmentation, presentation attack detection, and recognition. We use an image-complexity guided network compression scheme, CC-Net~\cite{Mishra_2019_CCNet}, to design a lightweight iris segmentation network that operates with high speed on Raspberry Pi, while producing high-quality masks. The presentation attack detection module is based on the best-performing methods among all available open-source iris PAD algorithms evaluated to date~\cite{fang2020robust}. For iris recognition, we employ an algorithm that extracts features using domain-specific, human-inspired Binarized Statistical Image Features (BSIF) \cite{Czajka2019_WACV_DomainSpecificBSIF} and performs matching with mean fractional Hamming Distance. \par

To summarize, the main {\bf contributions of this paper} are the following:
\begin{enumerate}
    \item An open source design of a complete iris recognition system equipped with iris presentation attack detection (hardware design based on Raspberry Pi and the end-to-end pipeline from image acquisition to recognition decision written in Python).
    \item A lightweight iris segmentation neural network that segments an ISO-compliant iris image in approx. 1.5 sec. on a Raspberry Pi 3B+, running around $3.5\times$ faster than the existing open-source iris segmentation (OSIRIS \cite{Othman_2016_OSIRIS}) and $38.2\times$ faster than the state-of-the-art deep learning-based iris segmentation (SegNet \cite{Trokielewicz_2018_IrisSegNet}) while preserving good segmentation accuracy.
\end{enumerate}

The complete package (segmentation model, source codes, hardware specifications and assembly instructions) can be obtained at \url{https://github.com/CVRL/RaspberryPiOpenSourceIris}. Since the implementation is done in Python, all software components can be easily run outside the Raspberry Pi environment and adapted to other needs.

\section{Related Work}

\paragraph{Open Source Biometric Systems.}

The idea of creating open-source biometric systems is not new, and has been developing unevenly across modalities due to uneven ubiquitousness of appropriate sensors. Among the three most popular biometric techniques (fingerprints, face, and iris), face recognition is the method that can be effectively implemented with commodity cameras, and thus several open-source software-only solutions exist \cite{openbr,Deng_CVPR_2019,Parkhi15}. Contactless fingerprint acquisition (with a camera installed in a mobile phone) has been also explored \cite{Kumar_2018}, which opens the door for low-cost/free fingerprint recognition systems. A contact-based open-source fingerprint recognition system has been recently proposed by Engelsma~\etal~\cite{Engelsma_2019_RaspiReader}. The authors built RaspiReader, an easy-to-assemble, spoof-resistant, optical fingerprint reader. RaspiReader provides open source STL and software which allow replication of the entire system within one hour at the cost of US \$175. Visible-light iris recognition has been also explored \cite{Proenca2016,Trokielewicz_JTIT_2016}, opening the possibilities for low-cost iris recognition solutions, although their recognition accuracy may be lower than that for near infrared-based systems, especially for dark eyes (\ie rich in melanin pigment).

\paragraph{Open Source Iris Recognition Software.}

There are several iris image processing and coding algorithms that have been open-sourced. Perhaps the first open-source re-implementation of Daugman's coding \cite{Daugman_1993_recognition} was published by Masek and Kovesi \cite{Masek}. It uses a bank of log-Gabor wavelets for feature extraction and the fractional Hamming distance for template matching. Masek's codes were later superseded by Open Source IRIS (OSIRIS) \cite{Othman_2016_OSIRIS}, written in C, which is arguably the most popular open source iris recognition software based on old concepts using circular approximations of iris boundaries and Gabor filters in iris encoding. University of Salzburg's Iris Toolkit (USIT) \cite{USIT3} extends the iris segmentation and coding beyond Gabor-based algorithm, and is the most recent open-source iris recognition software.

Iris recognition typically involves two major steps: (a) image segmentation and (b) feature extraction and comparison. These steps are in principle independent, and thus we observe open-source solutions for either segmentation or iris coding. Most initial works on iris segmentation focused on circular approximations of the inner and outer iris boundaries~\cite{Daugman_1993_recognition}. Since then, solutions of iris segmentation deal with more complex shapes, enabled by a recent advent of deep learning. Especially convolutional neural networks (CNN) have been used as end-to-end, effective iris image segmenters with model weights offered along with the papers \cite{Jalilian_2017_FCNSeg,kerrigan2019OTSDNN,Lozej_IWOBI_2018,Trokielewicz_2018_IrisSegNet}. Kinnison \etal \cite{Kinnison_ICB_2019} adapt the existing learning-free method designed earlier for neural microscopy volumes to iris segmentation and achieved considerable processing speed-up at low-resource platforms (such as Raspberry Pi) at a small cost of lower segmentation accuracy. Once the segmentation stage is secured, one may find open-sourced feature encoding methods based on standard \cite{Raja_IWBF_2014} and human-driven Binary Statistical Image Features \cite{Czajka2019_WACV_DomainSpecificBSIF}, or off-the-shelf CNN embeddings \cite{boyd2020deepIrisRec,Nguyen_Access_2018}. Ahmad and Fuller~\cite{ahmad2019thirdeye} propose a triplet CNN-based iris recognition system without a need for image segmentation and normalization of iris images.

Despite richness of proposed iris recognition methods, with some of them offering open-source codes, to the best of our knowledge no open-source hardware with appropriate codes suited for low-resource platforms, has been proposed yet for iris biometrics to date.


\paragraph{Open Source Iris Presentation Attack Detection (PAD) Software.}

For iris recognition to be effective and meaningful, presentation attacks (such as printed images, textured contact lenses, and even cadaver irises) to iris sensors must be detected and excluded from the matching process. The 2018 survey~\cite{Czajka_2018_PADsurvey} provides a comprehensive study of iris PAD research, covering background, presentation attack instruments, detection methods, datasets, and competitions. We will focus on the works most closely related to the topic of this paper, \ie only those iris PAD methods that have been open sourced. Hu \etal~\cite{Hu_2016_RegionalFeatures} use spatial pyramids and relational measures (convolutions on features) to extract higher-level regional features from local features extracted by Local Binary Patterns (LBP), Local Phase Quantization (LPQ), and intensity correlogram. Gragnaniello~\etal~\cite{Gragnaniello_2016_IrisSclera} extract features from both iris and sclera using scale-invariant local descriptors (SID) and apply a linear Support Vector Machine (SVM) for PAD. In another work by Gragnaniello~\etal~\cite{Gragnaniello_2016_IrisSclera}, the authors combine domain-specific knowledge into the design of their CNN, namely the network architecture and loss function. The PAD methods we adopt in this paper are the most recent, open-source methods evaluated by Fang \etal \cite{fang2020robust} as best-performing in their experiments: OSPAD-2D~\cite{McGrath_2018_OSPAD2D}, OSPAD-3D~\cite{Czajka_2019_OSPAD3D}, and OSPAD-fusion~\cite{fang2020robust}. OSPAD-2D is a multi-scale BSIF-based PAD method on textural features, OSPAD-3D is a photometric-stereo-based PAD method on shape features, and OSPAD-fusion is a combination of them. These methods will be discussed in further detail in Section~\ref{sec:software_PAD}.


\section{Design}

\subsection{Software}

\subsubsection{Iris Segmentation}
\label{sec:software_segmentation}

The most recent and most accurate iris segmentation methods deploy various convolutional neural network architectures. All of them, however, require more computing power and memory than a device like Raspberry Pi could offer. Therefore, we adapted the lightweight CC-Net architecture~\cite{Mishra_2019_CCNet} to iris segmentation. CC-Net has a U-Net structure~\cite{Ronneberger_2015_UNet}, but it leverages information about the image complexity of the training data to compress the network while preserving most of the segmentation accuracy. In the original paper, the authors showed that CC-Net is able to retain up to 95\% accuracy using only 0.1\% of the trainable parameters on biomedical image segmentation tasks. We applied the CC-Net network compression scheme to the domain of iris segmentation and ended up with a deep learning-based segmentation model that completes the prediction for ISO-compliant iris image in about 1.5 seconds on Raspberry Pi 3B+. That is, it runs around $3.5\times$ faster than the existing open-source iris segmentation (OSIRIS \cite{Othman_2016_OSIRIS}) and $38.2\times$ faster than the state-of-the-art deep learning-based segmentation (SegNet \cite{Trokielewicz_2018_IrisSegNet}), which is re-trained from an off-the-shelf SegNet model~\cite{Badrinarayanan_2017_SegNet} on the iris segmentation task. SegNet obtains benchmark performance and produces masks with fine details. However, it has a deep encoder-decoder structure and has a lot more parameters than CC-Net, making it a less ideal candidate for applications on mobile devices with limited memory and computing resource. In Section~\ref{sec:timings}, we compare the running time of all three iris segmentation methods on the Raspberry Pi, and verify our choice of CC-Net as the segmentation tool. Our software implementation includes also both SegNet-based and OSIRIS-based codes prepared for the Pi platform, to make the results presented in this paper reproducible. 


\subsubsection{Iris Recognition}

We adopt the domain-specific human-inspired BSIF-based iris recognition algorithm \cite{Czajka2019_WACV_DomainSpecificBSIF}. In contrast to previous work that treated BSIF as a generic textural filter, Czajka~\etal~\cite{Czajka2019_WACV_DomainSpecificBSIF} perform domain adaptation to train BSIF filters specifically learned for iris recognition. Additionally, the filters are trained on iris regions that appear important to humans to preserve the maximum amount of information possible. To further ensure the quality of images, only patches from genuine iris pairs that were correctly classified by humans are used in the filter training. The human-in-the-loop design and domain adaptation result in a statistically significant improvement in iris recognition accuracy, including methods based on standard BSIF filters and OSIRIS implementation of Gabor-based encoding, according to \cite{Czajka2019_WACV_DomainSpecificBSIF}.

\subsubsection{Iris Presentation Attack Detection}
\label{sec:software_PAD}

Despite many iris presentation attack detection methods proposed to date \cite{Czajka_2018_PADsurvey}, a small number of them have been formally open-sourced. Fang \etal \cite{fang2020robust} offer a recent comparison of available open-source iris PAD methods, and based on their assessment we decided to adopt the best-performing methods (OSPAD-3D \cite{Czajka_2019_OSPAD3D}, OSPAD-2D \cite{McGrath_2018_OSPAD2D}, and fusion of these two) and prepare a faster implementation by parallelization of processes and early stopping (for OSPAD-2D), suitable for low-resource platforms. Below, we provide a brief overview of each method and direct the readers to the original papers and our code implementation for more details.


\paragraph{OSPAD-3D} In OSPAD-3D, Czajka~\etal~\cite{Czajka_2019_OSPAD3D} leverage the fact that when iris is illuminated from two different directions, differences in shadows observed in the images of authentic iris are small, while significant differences appear in the images of irises wearing a textured contact lens. From this observation, OSPAD-3D employs photometric stereo to reconstruct the iris surface, which is flatter for authentic irises and more irregular for textured contact lenses. Given a pair of iris images and their segmentation masks, OSPAD-3D first estimates the surface normal vectors of the iris surface. Then, the variance of the vectors' distances to the mean normal vector is computed as the PAD score.

\paragraph{OSPAD-2D} Extending~\cite{Doyle_2015_BSIFPAD}, OSPAD-2D~\cite{McGrath_2018_OSPAD2D} proposes a PAD method using purely open source libraries. In this method, features are extracted through multi-scale standard BSIF, and an ensemble of Support Vector Machine, Multi-layer Perceptron, and Random Forests classifiers is used to make the final prediction. The method does not use iris segmentation and instead uses the best guess of iris location (center of the image) to obtain the salient iris region. Since OSPAD-2D integrates majority voting of the SVM decisions, we implement an early stopping strategy to further speed up the processing time: for each sample, once more than half of the models agree on a decision (live or fake), the decision is returned immediately and the rest of the models are not run.

\paragraph{OSPAD-fusion} Since OSPAD-3D uses only 3D shape features and OSPAD-2D uses only 2D textural features, Fang~\etal~\cite{fang2020robust} propose OSPAD-fusion that fuses the previous two methods. In the paper, the authors observe that lenses with highly opaque patterns often fail to produce well-pronounced shadows, thus violating the assumptions of OSPAD-3D. It is also observed that OSPAD-2D tends to accept fakes when tested on unknown samples. Therefore, instead of a simple score-level fusion, OSPAD-fusion employs a cascaded fusion algorithm that is able to sidestep the flaws present in OSPAD-3D and combine the strengths of the two methods.

\subsection{Hardware} 


Figure~\ref{fig:setup} (a-b) shows the setup of our hardware. The design is simple: an IR-sensitive camera (with the IR filter) is used to capture the iris images, and a circuit controlled by the Raspberry Pi output pins is used for illumination purposes. Since OSPAD-3D requires two iris images taken when the eye is illuminated from two different directions, two near-infrared LEDs (with a power peak around 850 nm) are placed equidistant to the lens and on both sides of the lens. At test time, the first picture is taken with the left LED turned on (and right LED turned off), and then the second picture is taken with the left LED turned off (and right LED turned on). Then, both LEDs will go off and the iris segmentation, recognition, and presentation attack detection steps are carried out.

\subsection{Assembly Details}
\begin{table}
\begin{center}
\caption{List of the main hardware components and their current costs.}
\label{table:hardware_price} 
\begin{tabular}{cc}
\toprule
{\bf Component} & {\bf Cost (USD)}\\ 
\midrule
Raspberry Pi 3B+ & \$$42.99$   \\ 
NIR-sensitive Pi-compatible & \\
camera ($2592 \times 1944$ px) & \$$13.99$ \\ 
NIR filter (cut at 760 nm) & \$$14.99$ \\ 
NIR LEDs (850 nm) & \$$0.20$ \\ 
Resistors (220 Ohm) & \$$1.04$ \\ 
\midrule
Total & \$$73.39$ \\ 
\bottomrule
\end{tabular}
\end{center}
\end{table}

Table~\ref{table:hardware_price} lists the main components used to construct the system and their current prices. The Raspberry Pi model version is 3B+. The NIR LEDs have emission wavelengths of 850mm, but any LEDs with wavelength in the range between 700mm and 900mm, as recommended by ISO \cite{ISO_19794_6_2011}, will work. Four 220 Ohm resistors are used for circuit protection purposes. For the software part, all software modules are implemented in Python. One may optionally use a protective metal sleeve, as shown in Fig.~\ref{fig:setup}, or skip using the NIR filter at all, if the camera has an embedded IR-cut filter. 

The complete package (segmentation model, source codes, hardware specifications, and assembly instructions) can be obtained at \url{https://github.com/CVRL/RaspberryPiOpenSourceIris}. Note that all software components can be easily run outside the Raspberry Pi environment, and thus serve as a rapid iris recognition implementation with presentation attack detection. The optional OSIRIS-based segmentation module (attached to the software release for comparisons and replication purposes) is written in C++.

\section{Performance Evaluation}

We evaluate the performance of iris recognition and presentation attack detection on \textit{NDIris3D} database \cite{fang2020robust}. This is the only database known to us which offers iris images taken from the same subjects with and without textured contact lenses, and captured user NIR illumination from two different directions (required for photometric stereo-based presentation attack detection method).

\subsection{Iris Recognition Performance}

This evaluation has two goals. One is obviously an assessment of the overall iris recognition accuracy offered by algorithms working in low-resource setup. The second goal is to compare the adapted, light-weight CC-Net-based segmentation with three other open-sourced segmenters: U-Net-based \cite{Lozej_2018_UNetSeg}, SegNet-based \cite{Trokielewicz_2018_IrisSegNet} and OSIRIS-based \cite{Othman_2016_OSIRIS}. In order to make a fair comparison, the iris images in the test set are first segmented by each segmentation method separately. Then, the same human-driven BSIF-based iris recognition algorithm~\cite{Czajka2019_WACV_DomainSpecificBSIF} is used to generate the genuine and imposter comparison scores using segmentation results of each segmentation method. It is noteworthy that all segmentation and iris coding methods used different, subject-disjoint datasets for training and hyperparameters setting, so this experimental protocol is the best available option. 


\begin{table}
\begin{center}
\caption{Iris recognition performance using masks computed by different segmentation methods.}
\label{table:d_prime_segmentation} 
\begin{tabular}{ccccc}
\toprule
\multirow{2}{*}{Method} & \multirow{2}{*}{$d'$}& EER & \multicolumn{2}{c}{FNMR(\%)}\\ \cline{4-5}
 & & (\%) & @FMR=1\% & @FMR=0.1\%\\ 
\midrule
CC-Net & $4.89$ & 4.39 & 8.48 & 19.36\\ 
OSIRIS & $4.90$ & 4.14 & 7.01 & 14.16\\ 
SegNet & $5.03$ & 4.26 & 8.63 & 18.75\\ 
U-Net & $0.82$ & 39.51 & 87.33 & 97.80\\ 
\bottomrule
\end{tabular}
\end{center}
\end{table}

\par 
The datasets used in training of both iris segmenters and the BSIF-based encoding method are subject-disjoint with the \textit{NDIris3D-LG} corpus (samples collected by the LG 4000 iris sensor) used for testing in this work. Table~\ref{table:d_prime_segmentation} reports the $d'$ between genuine and imposter distributions, Equal Error Rate (EER), and False Non-Match Rate (FNMR) at different levels of False Match Rate (FMR). As it can be seen, CC-Net, OSIRIS, and SegNet all achieve high $d'$ and low EER, which means that they give high-quality masks. Note that CC-Net achieves similar FMR as the state-of-the-art CNN-based segmentation method (SegNet). U-Net-based segmentation results in lackluster overall accuracy. This may be due to the fact that the CASIA-V1 dataset is small in size, and fine-tuning the U-Net weights with only 160 images (according to the original paper) may be insufficient to generalize on non-CASIA and ISO-compliant samples. Although it is possible to re-train it on a larger dataset, which would certainly lead to better performance, we will show in Section~\ref{sec:timings} that the structure of this network is prohibitively large to be deployable on low-resource platforms, such as Raspberry Pi.

\par
\begin{figure*}[!ht]
   \centering
   \includegraphics[width=0.18\textwidth]{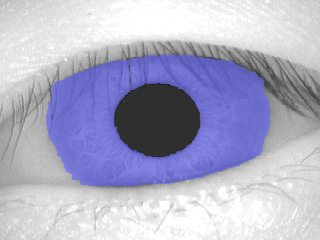}\quad
   \includegraphics[width=0.18\textwidth]{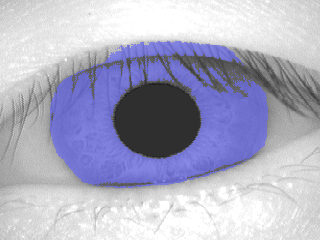} \quad
   \includegraphics[width=0.18\textwidth]{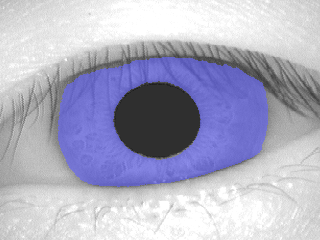}\quad
   \includegraphics[width=0.18\textwidth]{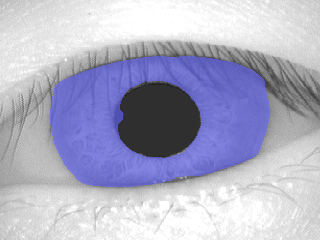}\quad
   \includegraphics[width=0.18\textwidth]{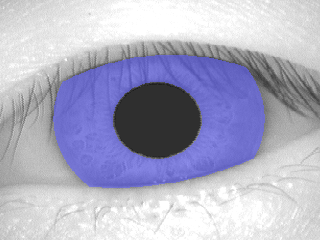}\\[-0.1ex]
   \includegraphics[width=0.18\textwidth]{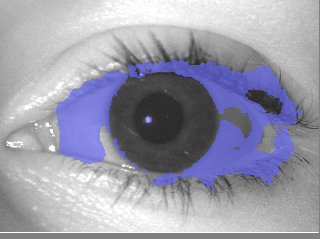}\quad
   \includegraphics[width=0.18\textwidth]{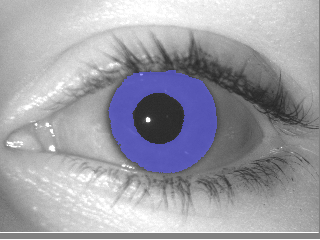} \quad
   \includegraphics[width=0.18\textwidth]{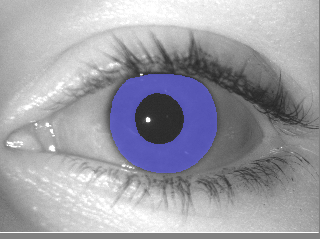}\quad
   \includegraphics[width=0.18\textwidth]{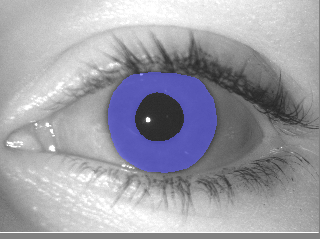}\quad
   \includegraphics[width=0.18\textwidth]{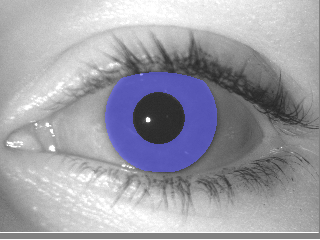}\\[-0.2ex]
   \subfigure[]{\includegraphics[width=0.18\textwidth]{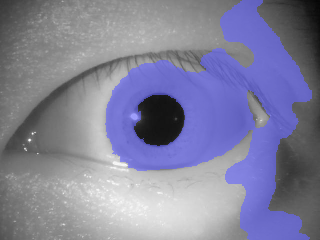}}\quad
   \subfigure[]{\includegraphics[width=0.18\textwidth]{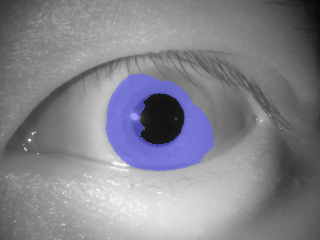}} \quad
   \subfigure[]{\includegraphics[width=0.18\textwidth]{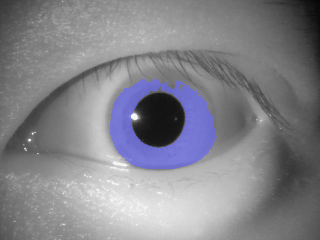}}\quad
   \subfigure[]{\includegraphics[width=0.18\textwidth]{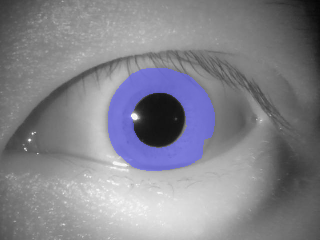}}\quad
   \subfigure[]{\includegraphics[width=0.18\textwidth]{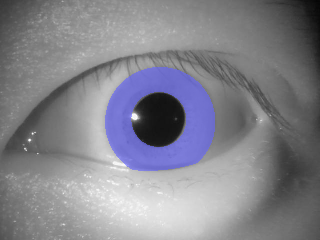}}\quad
   \caption{Examples of segmentation results for a sample from CASIA-V1 (\textbf{upper row}) and \textit{NDIris3D} (\textbf{middle row}), as well as an image captured by Raspberry Pi (\textbf{bottom row}), generated by: (a) U-Net~\cite{Lozej_2018_UNetSeg}, (b) OSIRIS~\cite{Othman_2016_OSIRIS}, (c) SegNet~\cite{Trokielewicz_2018_IrisSegNet}, (d) CC-Net~\cite{Mishra_2019_CCNet}, (e) human-labeled ground truth.}
   \label{fig:segm_performance}
\end{figure*}

We also present selected qualitative results of the segmentation results of different methods. Figure~\ref{fig:segm_performance} shows two examples from CASIA-V1 and \textit{NDIris3D-LG}. It can be seen that U-Net~\cite{Lozej_2018_UNetSeg} performs perfectly on CASIA-V1, but fails to generalize onto \textit{NDIris3D-LG}. OSIRIS misclassified part of the iris and sclera region for CASIA-V1 but achieves a robust performance overall. SegNet- and CC-Net-based methods generate masks that are close to the ground truth. Note that SegNet predicts finer details for the CASIA-V1 when compared to CC-Net segmentation, and we conjecture that this is because SegNet is a richer model, trained on a variety of data, and is therefore able to capture more low-level features. Both SegNet and CC-Net demonstrate very good segmentation results for iris images captured by the hardware proposed in this work. These qualitative observations correspond with our quantitative assessment.

\subsection{Iris Presentation Attack Detection Performance}

\begin{figure}[!htb]
\centering
\centering
\includegraphics[width=0.22\textwidth]{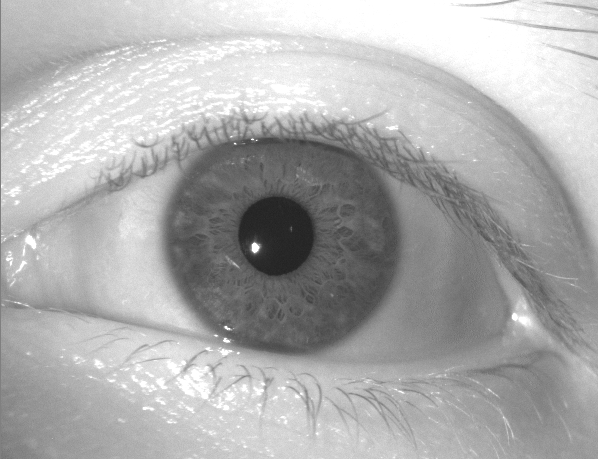}\quad
\includegraphics[width=0.22\textwidth]{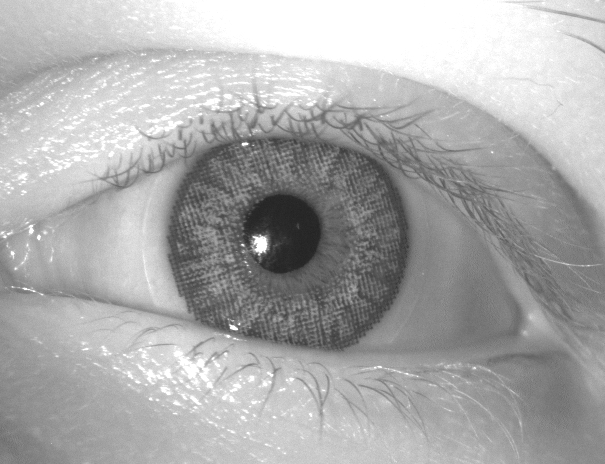}
\caption{Examples of images of a real iris (\textbf{left}) and an iris with textured contact lens (\textbf{right}) of the same subject. Part of the real iris texture is revealed even with the contact lens.}
\label{fig:real_contact}
\end{figure}

\begin{table}
\begin{center}
\caption{$d'$ for each pair of distributions.}
\label{table:d_prime_recognition} 
\begin{tabular}{cccc}
\toprule
Distribution pairs & $d'$\\ 
\midrule
Genuine vs Imposter & $4.887$\\ 
Genuine vs Real-Contact & $3.557$\\ 
Real-Contact vs Imposter & $1.445$\\ 
\bottomrule
\end{tabular}
\end{center}
\end{table}

\begin{figure}[!htb]
\centering
\includegraphics[width=0.5\textwidth]{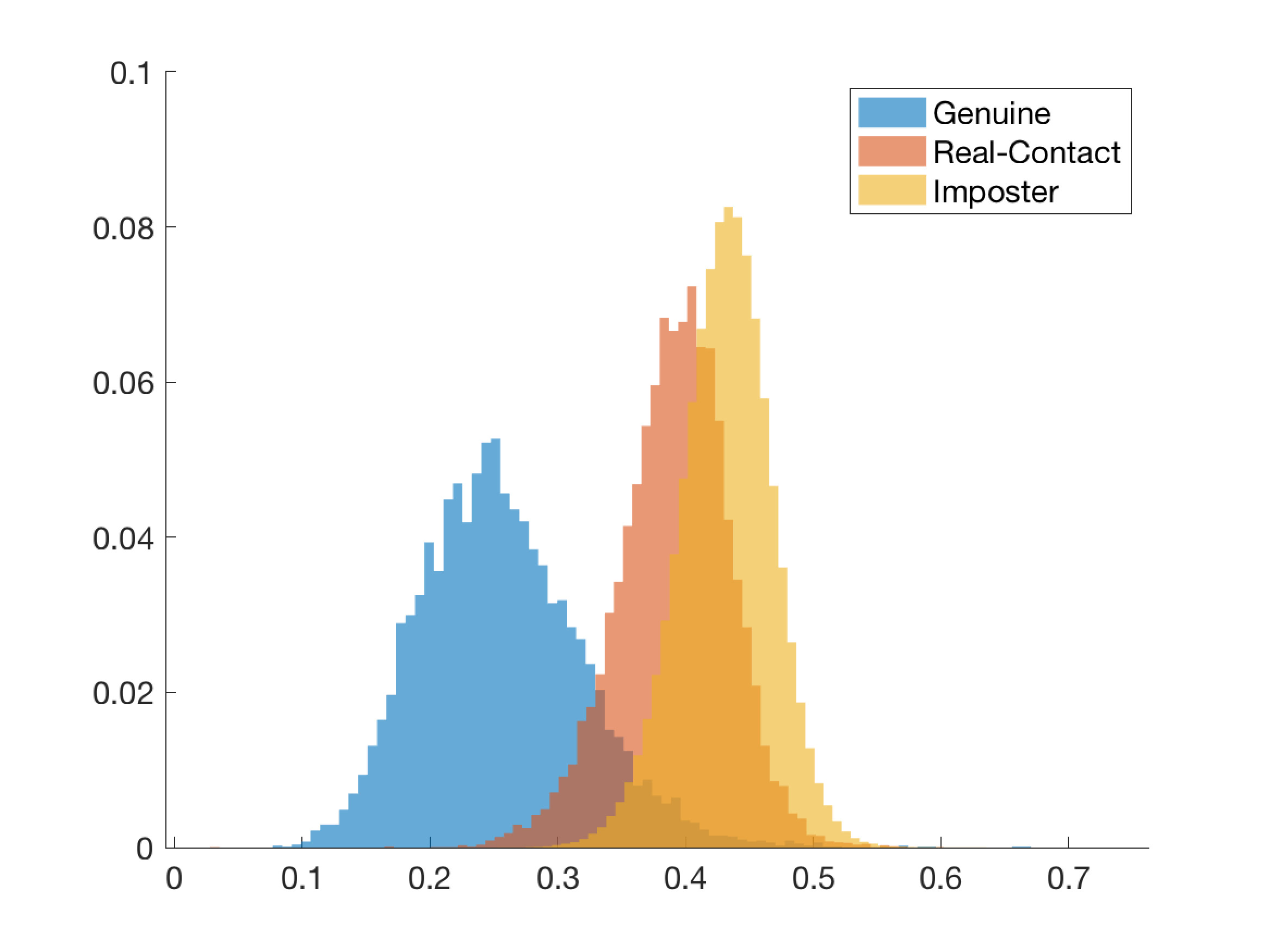}
\caption{Probability density (Y axis) for comparison scores (X axis): {\sf Genuine} (the same irises without contact lenses are compared), {\sf Real-Contact} (the same irises with and without contact lenses are compared), and {\sf Imposter} comparisons (different eyes are compared). The lower the score, the better the match.}
\label{fig:rec_dist}
\end{figure}

Placing textured contact lenses on the cornea to interfere with actual iris pattern, and thus conceal one's identity, is a well-known presentation attack type. A close observation of images may reveal that textured contact lenses cover only part of the iris, leaving some regions of the iris observable, as shown in Figure~\ref{fig:real_contact}. This fractional authentic iris area is, however, too small to obtain reliable recognition. Indeed, Table~\ref{table:d_prime_recognition} and Figure~\ref{fig:rec_dist} demonstrate that when the genuine irises are covered by textured contact lenses, the genuine score distribution (denoted as ``Genuine vs Real-Contact'') is shifted towards imposter score distribution, significantly increasing the probability of false non-match. Such attempts should be detected, and thus presentation attack detection is necessary. 

\begin{table*}
\begin{center}
\caption{Comparison of performance of the adopted iris PAD method with other available open-source iris PAD methods on the \textit{NDIris3D-LG} test database. BPCER $=$ Bona-fide Presentation Classification Error Rate (that is, the percentage of live irises called spoof). APCER $=$ Attack Presentation Classification Error Rate (that is, the percentage of spoofed irises called live).}
\label{table:PAD_performance} 
\begin{tabular}{ccccccc}
\toprule
\multirow{3}{*}{Method} & \multicolumn{6}{c}{Performance}\\
 & \multicolumn{3}{c}{LG4000} & \multicolumn{3}{c}{AD100} \\
\cline{2-7}
 & Accuracy (\%) & APCER (\%) & BPCER (\%) & Accuracy (\%) & APCER (\%) & BPCER (\%) \\
\midrule
OSPAD-fusion~\cite{fang2020robust} & $94.76$ & $6.36$ & $4.12$ & $91.14$ & $11.84$ & $5.92$ \\ 
OSPAD-3D~\cite{Czajka_2019_OSPAD3D} & $84.75$ & $3.89$ & $26.71$ & $80.43$ & $34.33$ & $4.97$ \\ 
OSPAD-2D~\cite{McGrath_2018_OSPAD2D} & $89.18$ & $21.27$ & $0.46$ & $79.54$ & $40.19$ & $0.95$ \\ 
DACNN~\cite{Gragnaniello_2016_DACNN} & $77.79$ & $29.84$ & $6.76$ & $59.19$ & $72.31$ & $8.97$ \\ 
SIDPAD~\cite{Gragnaniello_2016_IrisSclera} & $80.61$ & $31.10$ & $3.90$ & $73.35$ & $49.76$ & $3.76$ \\ 
RegionalPAD~\cite{Hu_2016_RegionalFeatures} & $78.37$ & $39.91$ & $3.44$ & $56.30$ & $64.24$ & $23.3$ \\ 
\bottomrule
\end{tabular}
\end{center}
\end{table*}

In this subsection, we compare the PAD methods implemented in our system (OSPAD-fusion, OSPAD-3D, and OSPAD-2D) to all currently available open-source baselines, by either following download instructions offered in original papers, or -- in some cases -- asking the corresponding authors for source codes. We evaluated all methods under the same protocol: all algorithms are trained according to the training schedule in the original papers / the schedules suggested by the authors, with the \textit{NDCLD'15} dataset \cite{Doyle_2015_BSIFPAD} as the training set, and the entire \textit{NDIris3D-LG} as the testing set. Again, our goal was to use a subject-disjoint train and test sets of images. Since OSPAD-3D generates one prediction for each pair of images, we use a validation subset from the training set \textit{NDCLD'15} to search for the best fusion strategy for the other methods. As a result, all methods now generate one prediction for each pair of images, so their performance could be evaluated fairly. Results in Table~\ref{table:PAD_performance} show that the adopted methods achieve state-of-the-art accuracy in iris presentation attack detection.

\subsection{Timings}
\label{sec:timings}

In order for an open-source system to be deployable in real-world application scenarios, the most important feature is fast running time. A slow system is not desirable, even if it is cost-effective, open software, and open hardware. Table~\ref{table:running_times} presents the average and standard deviation of the running times (in seconds) of each module in our system, on both \textit{NDCLD'15} and \textit{NDIris3D-LG} sets. Note that all the times are for a \textbf{pair} of iris images, as required by the OSPAD-3D method. So processing times per a single iris image are approximately half of those reported in Table \ref{table:running_times}. All modules have fast running time because our implementation takes advantage of all four cores of the Raspberry Pi through multiprocessing. We can further parallelize OSPAD-2D with the other modules, since it does not depend on segmentation results. Note that while the running times for the other modules are roughly comparable for the two datasets, both the average and standard deviation of the OSPAD-2D running time are significantly higher on \textit{NDIris3D-LG} than on \textit{NDCLD'15}. This is because the training set of OSPAD-2D models partially overlaps with our subset of \textit{NDCLD'15}. This makes OSPAD-2D decisions on \textit{NDCLD'15} more confident, thus triggering the early stopping more frequently on \textit{NDCLD'15} than on \textit{NDIris3D-LG}. 

\begin{table}
\begin{center}
\caption{Running times (seconds) of all components of the proposed system on \textit{NDCLD'15} and \textit{NDIris3D}. Standard deviation is shown in round brackets.}
\label{table:running_times} 
\vskip2mm
\begin{tabular}{lcc}
\toprule
Operation & \textit{NDCLD'15} & \textit{NDIris3D}\\ 
\midrule
Segmentation & $1.458~(\pm 0.054)$ & $1.488~(\pm 0.052)$  \\ 
OSPAD-3D & $0.186~(\pm 0.023)$ & $0.177~(\pm 0.020)$ \\ 
OSPAD-2D & $4.327~(\pm 0.489)$ & $5.889~(\pm 1.104)$ \\ 
Iris Recognition & $1.735~(\pm 0.019)$ & $1.876~(\pm 0.290)$  \\ 
\bottomrule
\end{tabular}
\end{center}
\end{table}

We also compare the running times of CC-Net, compared to the other three candidate segmentation tools: OSIRIS segmentation module, SegNet-based and U-Net-based. Table~\ref{table:segm_times} shows the running times of each method, along with the number of trainable parameters. Results show that CC-Net is the fastest segmentation method, around $3.5\times$ faster than OSIRIS and $38.2\times$ faster than SegNet. The running time of U-Net is not reported because the model is too large to be run on Raspberry Pi. The huge speed advantage of CC-Net is due to the fact that the number of its trainable parameters is significantly less than SegNet and U-Net. Though OSIRIS does not need to tune any parameters, it requires a lot of computation to perform the Viterbi algorithm. Therefore, CC-Net is left as the segmentation component in the final prototype.

\begin{table}
\begin{center}
\caption{Running times (seconds) measured on Raspberry Pi for three segmentation methods on the \textit{NDCLD'15} dataset.}
\label{table:segm_times} 
\vskip2mm
\begin{tabular}{lcc}
\toprule
Method & Running time & \# of trainable parameters\\ 
\midrule
CC-Net & $1.458~(\pm 0.054)$ & $1.2e5$  \\ 
OSIRIS & $5.224~(\pm 0.639)$ & $0$ \\ 
SegNet & $55.637~(\pm 2.197)$ & $2.9e7$\\ 
UNet & $-$ & $1.2e8$\\ 
\bottomrule
\end{tabular}
\end{center}
\end{table}

\section{Summary}

This paper proposes the first, known to us, complete open-source iris recognition system (hardware and software) that can be implemented on low-resource platforms supporting Python, such as Raspberry Pi. It features not only a fast and accurate CNN-based segmentation (which is usually the most time-consuming element of iris recognition pipeline) but also one of the most recent iris presentation attack detection methods, making it complete also in terms of security aspects. It may serve as a state-of-the-art open source platform for designing secure and low-cost iris recognition systems. All software components can be also easily run outside the Raspberry Pi, and even adapted to other hardware platforms supporting Python programming language and near-infrared imaging.

{\small
\bibliographystyle{ieee}
\bibliography{submission_example}
}

\end{document}